\documentclass[11pt,a4paper]{article}
\usepackage[hyperref]{emnlp-ijcnlp-2019}
\usepackage{times}
\usepackage{latexsym}

\usepackage{amsmath,amssymb}
\usepackage{algorithmic}
\usepackage{graphicx,subcaption}
\usepackage{textcomp}
\usepackage{xcolor}
\usepackage[ruled]{algorithm2e}

\usepackage{url}

\DeclareMathOperator{\E}{\mathbb{E}}

\aclfinalcopy % Uncomment this line for the final submission
\title{Neural Gaussian Copula for Variational Autoencoder}

\author{Prince Zizhuang Wang \\
  Department of Computer Science \\
  University of California Santa Barbara \\
  \texttt{zizhuang\textunderscore wang@ucsb.edu} \\\And
  William Yang Wang \\
  Department of Computer Science \\
  University of California Santa Barbara \\
  \texttt{william@cs.ucsb.edu}
  }

\date{}

\begin{document}
\maketitle
\begin{abstract}
   Variational language models seek to estimate the posterior of latent variables with an approximated variational posterior. The model often assumes the variational posterior to be factorized even when the true posterior is not. The learned variational posterior under this assumption does not capture the dependency relationships over latent variables. We argue that this would cause a typical training problem called posterior collapse observed in all other variational language models. We propose Gaussian Copula Variational Autoencoder (VAE) to avert this problem. Copula is widely used to model correlation and dependencies of high-dimensional random variables, and therefore it is helpful to maintain the dependency relationships that are lost in VAE. The empirical results show that by modeling the correlation of latent variables explicitly using a neural parametric copula, we can avert this training difficulty while getting competitive results among all other VAE approaches. \footnote{Code will be released at \url{https://github.com/kingofspace0wzz/copula-vae-lm}}
\end{abstract}

\section{Introduction}
\textbf{Variational Inference (VI)}~\cite{wainwright2008graphical, hoffman2013stochastic} methods are inspired by \textit{calculus of variation}~\cite{gelfand2000calculus}. It can be dated back to the 18th century when it was mainly used to study the problem of the change of \textit{functional}, which is defined as the mapping from functions to real space and can be understood as the function of functions. \textbf{VI} takes a distribution as a \textit{functional} and then studies the problem of matching this distribution to a target distribution using \textit{calculus of variation}. After the rise of \textit{deep learning}~\cite{krizhevsky2012imagenet}, a deep generative model called \textbf{Variational Autoencoder}~\cite{kingma2014auto, hoffman2013stochastic} is proposed based on the theory of \textbf{VI} and achieves great success over a huge number of tasks, such as transfer learning~\cite{shen2017style}, unsupervised learning~\cite{jang2016categorical}, image generation~\cite{gregor2015draw}, semi-supervised classification~\cite{jang2016categorical}, and dialogue generation~\cite{zhao2017learning}. VAE is able to learn a continuous space of latent random variables which are useful for a lot of classification and generation tasks.

\begin{figure}[t]
\small
    \centering
    \includegraphics[width=0.315\textwidth]{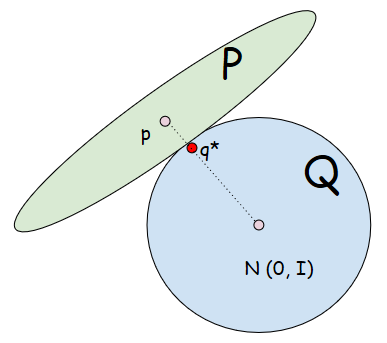}
    \caption{Intuitive illustration of \textbf{VI}: The elliptic $P$ is a distribution family containing the true posterior $p \in P$, and the circle $Q$ is a Mean-field variational family containing a standard normal prior $\mathcal{N}$. The optimal solution $q^*$ is the one in $Q$ that has the smallest $KL(q||p)$. In reality these two families may not overlap.}
    \label{fig:vi}
    \vspace{-3ex}
\end{figure}

Recent studies~\cite{bowman2015generating, yang2017improved, xiao2018dirichlet, xu2018spherical} show that when it comes to text generation and language modeling, VAE does not perform well and often generates random texts without making good use of the learned latent codes. This phenomenon is called \textbf{Posterior Collapse}, where the Kullback-Leibler (KL) divergence between the posterior and the prior (\textit{often assumed to be a standard Gaussian}) vanishes. It makes the latent codes completely useless because any text input will be mapped to a standard Gaussian variable. Many recent studies~\cite{yang2017improved, xu2018spherical, xiao2018dirichlet, miao2016neural, he2018lagging} try to address this issue by providing new model architectures or by changing the objective functions. Our research lies in this second direction. \textbf{We review the theory of VAE, and we argue that one of the most widely used assumptions in VAE, the Mean-field assumption, is problematic}. It assumes that all approximated solutions in a family of variational distributions should be factorized or dimensional-wise independent for tractability. We argue that it leads to the posterior collapse problem since any variational posterior learned in this way does not maintain the correlation among latent codes and will never match the true posterior which is unlikely factorized.

We avert this problem by proposing a \textbf{Neural Gaussian Copula (Copula-VAE)} model to train VAE on text data. Copula~\cite{nelsen2007introduction} can model dependencies of high-dimensional random variables and is very successful in risk management~\cite{kole2007selecting, mcneil2005quantitative}, financial management~\cite{wang2014semiparametric}, and other tasks that require the modeling of dependencies. We provide a reparameterization trick~\cite{kingma2014auto} to incorporate it with VAE for language modeling. We argue that by maintaining the dependency relationships over latent codes, we can dramatically improve the performance of variational language modeling and avoid posterior collapse. Our major contributions can be summarized as the following:
\begin{itemize}
  \item We propose Neural parameterized Gaussian Copula to get a better estimation of the posterior for latent codes.  
   \item We provide a reparameterization technique for Gaussian Copula VAE. The experiments show that our method achieves competitive results among all other variational language modeling approaches.
  \item We perform a thorough analysis of the original VAE and copula VAE. The results and analysis reveal the salient drawbacks of VAE and explain how introducing a copula model could help avert the posterior collapse problem.
\end{itemize}

\section{Related Work}

\paragraph{Copula: Before the rise of Deep Learning}
Copula~\cite{nelsen2007introduction} is a multivariate distribution whose marginals are all uniformly distributed. Over the years, it is widely used to extract correlation within high-dimensional random variables, and achieves great success in many subjects such as risk management~\cite{kole2007selecting, mcneil2005quantitative}, finance~\cite{wang2014semiparametric}, civil engineering\cite{chen2012drought, zhang2006bivariate}, and visual description generation~\cite{WangWen:2015}. In the past, copula is often estimated by Maximum Likelihood method~\cite{choros2010copula, jaworski2010copula} via parametric or semi-parametric approaches~\cite{tsukahara2005semiparametric, choros2010copula}. One major difficulty when estimating the copula and extracting dependencies is the dimensionality of random variables. To overcome the curse of dimensionality, a graphical model called \textit{vine copula}~\cite{joe2011dependence, czado2010pair, bedford2002vines} is proposed to estimate a high-dimensional copula density by breaking it into a set of bivariate conditional copula densities. However, this approach is often hand-designed which requires human experts to define the form of each bivariate copula, and hence often results in overfitting. Therefore, Gaussian copula~\cite{xue2000multivariate, frey2001copulas} is often used since its multivariate expression has a simple form and hence does not suffer from the curse of dimensionality.

\paragraph{VAE for Text}
~\citet{bowman2015generating} proposed to use VAE for text generation by using LSTM as encoder-decoder. The encoder maps the hidden states to a set of latent variables, which are further used to generate sentences. While achieving relatively low sample perplexity and being able to generate easy-to-read texts, the LSTM VAE often results in posterior collapse, where the learned latent codes become useless for text generation.

Recently, many studies are focusing on how to avert this training problem. They either propose a new model architecture or modify the VAE objective. ~\citet{yang2017improved} seeks to replace the LSTM~\cite{hochreiter1997long} decoder with a CNN decoder to control model expressiveness, as they suspect that the over-expressive LSTM is one reason that makes KL vanish. ~\citet{xiao2018dirichlet} introduces a topic variable and pre-trains a \textit{Latent Dirichlet Allocation}~\cite{blei2003latent} model to get a prior distribution over the topic information. ~\citet{xu2018spherical} believes the \textit{bubble soup effect} of high-dimensional Gaussian distribution is the main reason that causes KL vanishing, and therefore learns a hyper-spherical posterior over the latent codes. 
% We believe the problem comes from the nature of variational family itself and hence we propose our \textbf{Copula-VAE} which make use of the dependency modeling ability of copula model to guide the variational posterior to match the true posterior. We will provide more details of variational inference and copula-VAE in the following sections.

\section{Variational Inference}
The problem of inference in probabilistic modeling is to estimate the posterior density $p(\bold z|\bold x)$ of latent variable $\bold z$ given input samples $\{x_i\}_{i=1}^D$. The direct computation of the posterior is intractable in most cases since the normalizing constant $\int p(\bold z,\bold x)d\bold z$ lacks an analytic form. To get an approximation of the posterior, many approaches use sampling methods such as Markov chain Monte Carlo (MCMC)~\cite{gilks1995markov} and Gibbs sampling~\cite{george1993variable}. The downside of sampling methods is that they are inefficient, and it is hard to tell how close the approximation is from the true posterior. The other popular inference approach, \textbf{variational inference (VI)}~\cite{wainwright2008graphical, hoffman2013stochastic}, does not have this shortcoming as it provides a distance metric to measure the fitness of an approximated solution. 

In \textbf{VI}, we assume a variational family of distributions $\mathcal{Q}$ to approximate the true posterior. The \textit{Kullback-Leibler} (\textbf{KL}) divergence is used to measure how close $q \in Q$ is to the true $p(\bold z|\bold x)$. The optimal variational posterior $q* \in Q$ is then the one that minimizes the KL divergence 
\begin{align*}
    KL(q||p) = \sum q(\bold z|\bold x) log\frac{ q(\bold z|\bold x)}{p(\bold z|\bold x)}
\end{align*}

Based on this, \textit{variational autoencoder} (\textbf{VAE})~\cite{kingma2014auto} is proposed as a latent generative model that seeks to learn a posterior of the latent codes by minimizing the KL divergence between the true joint density $p_\theta(\bold x,\bold z)$ the variational joint density $q_\phi(\bold z,\bold x)$. This is equivalent to maximizing the following \textit{evidence lower bound} \textbf{ELBO},
\begin{align*}
\mathcal{L}(\bold \theta; \bold \phi; \bold x) & = -KL(q_\phi(\bold z,\bold x)||p_\theta(\bold x,\bold z))\\
& = \E_{q(\bold x)}[\E_{q_\phi(\bold z| \bold x)}[\log p_\theta(\bold x| \bold z)] \\
& \qquad - KL(q_\phi(\bold z|\bold x)||p(\bold z))]
\end{align*}

In this case, Mean-field~\cite{kingma2014auto} assumption is often used for simplicity. That is, we assume that the members of variational family $Q$ are dimensional-wise independent, meaning that the posterior $q$ can be written as $q(\bold z|\bold x) = \prod_{i=1}^D q(z_i|\bold x)$. The simplicity of this form makes the estimation of ELBO very easy. However, it also leads to a particular training difficulty called posterior collapse, where the KL divergence term becomes zero and the factorized variational posterior collapses to the prior. The latent codes $\bold z$ would then become useless since the generative model $p(\bold x|\bold z)$ no longer depends on it.

We believe the problem comes from the nature of variational family itself and hence we propose our \textbf{Copula-VAE} which makes use of the dependency modeling ability of copula model to guide the variational posterior to match the true posterior. We will provide more details in the following sections.

We hypothesize that the Mean-field assumption is problematic itself as the $q$ under this assumption can never recover the true structure of $p$. On the other hand, \textbf{Copula-VAE} makes use of the dependency relationships maintained by a copula model to guide the variational posterior to match the true posterior. Our approach differs from \textit{Copula-VI}~\cite{tran2015copula} and \textit{Gaussian Copula-VAE}~\cite{suh2016gaussian} in that we use copula to estimate the joint density $p(\bold z)$ rather than the empirical data density $p(\bold x)$. 

\section{Our Approach: Neural Gaussian Copula}

\subsection{Gaussian Copula}
In this section, we review the basic concepts of Gaussian copula. Copula is defined as a probability distribution over a high-dimensional unit cube $[0,1]^d$ whose univariate marginal distributions are uniform on $[0,1]$. Formally, given a set of uniformaly distributed random variables $U_1, U_2,...,U_n$, a copula is a joint distribution defined as
\begin{align*}
C(u_1, u_2,...,u_n) = P(U_1 \le u_1,..., U_n\le u_n)
\end{align*}

What makes a copula model above so useful is the famous \textbf{Sklar's Theorem}. It states that for any joint cumulative distribution function \textbf{(CDF)} with a set of random variables $\{ x_i\}_1^d$ and marginal CDF $F_i(x_i) = P(X_i \le x)$, there exists one unique copula function such that the joint CDF is
\begin{align*}
    F(x_1,...,x_d) = C(F_1(x_1),...,F_d(x_d))
\end{align*}
By \textit{probability integral transform}, each marginal CDF is a uniform random variable on $[0,1]$. Hence, the above copula is a valid one. Since for each joint CDF, there is one unique copula function associated with it given a set of marginals, we can easily construct any joint distribution whose marginal univariate distributions are the ones $F_i(x_i)$ that are given. And, for a given joint distribution, we can also find the corresponding copula which is the CDF function of the given marginals.

A useful representation we can get by \textbf{Sklar's Theorem} for a continuous copula is,
\begin{align*}
    C(u_1,...u_d) & = F(F_1^-(u_1),...,F_d^-(u_d))
\end{align*}

If we further restrict the marginals to be Gaussian, then we can get an expression for Gaussian copula, that is,
\begin{align*}
    C_\Sigma (u_1,...,u_d) = \Phi(\Phi_1^-(u_1),...,\Phi_d^-(u_d);0, \Sigma)
\end{align*}
where $\Phi(\cdot;\Sigma)$ is the CDF of a multivariate Gaussian Distribution $\mathcal{N}(0, \Sigma)$, and $\{\Phi_i^{-}\}$ is the inverse of a set of marginal Gaussian CDF.

To calculate the joint density of a copula function, we take the derivative with respect to random variables $u$ and get
\begin{align*}
    c_\Sigma(u_1,...,u_d) & = \frac{\partial C_\Sigma(u_1,...,u_d)}{\partial u_1\cdots u_d}\\
    & = \frac{\partial C_\Sigma(u_1,...,u_d)}{\partial q_1\cdots q_d}\prod^d_{i=1} \frac{\partial q_i}{\partial u_i}\\
    & = (\prod_{i=1}^d\sigma_i)|\Sigma^{-1/2}|exp(-\frac{1}{2} q^TMq)
\end{align*}
where $M = \Sigma^- - diag(\Sigma)^-$, and $q_i = \Phi_i^-(u_i)$.

Then, if the joint density $p(x_1,...,x_d)$ has a Gaussian form, it can be expressed by a copula density and its marginal densities, that is,
\begin{align*}
    p(x_1,...,x_d) & = \frac{\partial F(\cdot)}{\partial x_1\cdots x_d}\\
    & = \frac{\partial C_\Sigma(\cdot)}{\partial u_1\cdots u_d}\prod \frac{\partial u_i}{\partial x_i}\\
    & = c_\Sigma(u_1,...,u_d)\prod_i p(x_i)
\end{align*}
Therefore, we can decompose the problem of estimating the joint density into two smaller sub-problems: one is the estimation for the marginals; the other is the estimation for the copula density function $c_\Sigma$. In many cases, we assume independence over random variables due to the intractability of the joint density. For example, in the case of variational inference, we apply Mean-Field assumption which requires the variational distribution family to have factorized form so that we can get a closed-form KL divergence with respect to the prior. This assumption, however, sacrifices the useful dependency relationships over the latent random variables and often leads to training difficulties such as the posterior collapse problem. If we assume the joint posterior of latent variables to be Gaussian, then the above Gaussian copula model can be used to recover the correlation among latent variables which helps obtain a better estimation of the joint posterior density. In the VAE setting, we can already model the marginal independent posterior of latent variables, so the only problem left is how to efficiently estimate the copula density function $c_\Sigma$.
% Traditionally, this copula density can be parameterized and then be estimated using Maximum Likelihood~\cite{choros2010copula, jaworski2010copula}, or it can be estimated using a semi-parametrized method by first constructing an empirical CDF function and then doing Rank-based estimation~\cite{tsukahara2005semiparametric}. 
In the next section, we introduce a neural parameterized Gaussian copula model, and we provide a way to incorporate it with the reparameterization technique used in VAE. 

\subsection{Neural Gaussian Copula for VI}

\begin{figure*}[t]
\small
    \begin{center}
        \includegraphics[width=.8\textwidth]{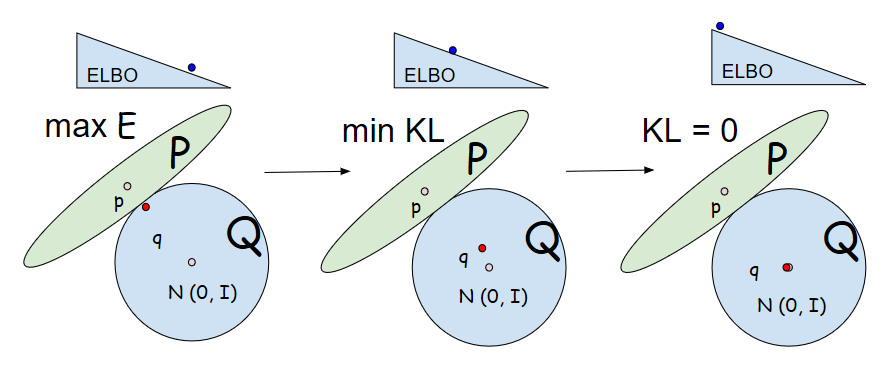}
    \end{center}
    \caption{Training stage of VAE. Initially, the model tries to maximize ELBO by maximizing $\E_{q(\bold z| \bold x)}[p(\bold x|\bold z)]$. Once $\E_{q(\bold z| \bold x)}[p(\bold x|\bold z)]$ is maximized, the model maximizes ELBO by minimizing KL. During this stage, the posterior starts to move closer to the prior. In the final stage, the posterior collapses to the prior. But, the ELBO and $\E_{q(\bold z| \bold x)}[p(\bold x|\bold z)]$ are already maximized, which means the model keeps constraining KL and there are not enough gradients to move the posterior away from the prior anymore.}
    \label{fig:stage}
\end{figure*}

\begin{figure}[t]
\small
\centering
    \includegraphics[width=.2\textwidth]{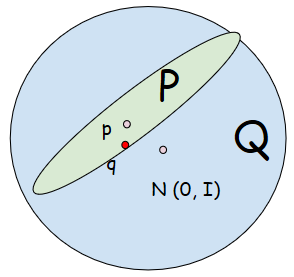}
    \caption{\textbf{Ideal} final stage of Copula-VAE. The family of distributions that contains the true posterior is now a subset of the variational family.}
    \label{fig:copula}
\end{figure}

By Mean-field assumption, we construct a variational family $\mathcal{Q}$ assuming that each member $q\in \mathcal{Q}$ can be factorized,
\begin{align*}
    q(\bold z) = \prod_i^d q(z_i)
\end{align*}

This assumption, however, loses dependencies over latent codes and hence does not consider the non-factorized form of the true posterior. In this case, as pictured by Figure~\ref{fig:stage}, when we search the optimal $q$*, it will never reach to the true posterior $p$. If we relieve the assumption, the variational family may overlap with the posterior family. However, this is intractable as the Monto Carlo estimator with respect to the objective often has very high variance~\cite{kingma2014auto}. Hence, we need to find a way to make it possible to match the variational posterior with the true posterior while having a simple and tractable objective function so that the gradient estimator of the expectation is simple and precise.

This is where Gaussian Copula comes into the story. Given a factorized posterior, we can construct a Gaussian copula for the joint posterior,
\begin{align*}
    q_\phi(\bold z|x) = c_\Sigma(\Phi_1(z_1),...\Phi(z_d))\prod_i^d q_\phi(z_i|x)
\end{align*}
where $c_\Sigma$ is the Gaussian copula density. If we take the log on both sides, then we have,
\begin{align*}
    \log q_\phi(\bold z|x) = \log c_\Sigma(u_1,...,u_d)+\sum_i^d \log q_\phi(z_i|x)
\end{align*}
Note that the second term on the right hand side is just the factorized log posterior we have in the original VAE model. By reparameterization trick~\cite{kingma2014auto}, latent codes sampled from the posterior are parameterized as a deterministic function of $\mu$ and $\sigma^2$, that is, $\bold z = \mu + \sigma \cdot \epsilon$, $\epsilon \sim \mathcal{N}(0, I)$, where $\mu, \sigma^2$ are parameterized by two neural networks whose inputs are the final hidden states of the LSTM encoder. Since $\prod_i q_\phi(z_i|x) = \mathcal{N}(\mu, \sigma^2I)$, we can compute the sum of log density of posterior by,
\begin{align*}
    \sum_i^d \log q_\phi(z_i|x) & = -\sum_i^d \log|\sigma_i| - \sum_i^d \frac{(z_i - \mu_i)^2}{2\sigma_i^2}\\
    & - \frac{d}{2}\log 2\pi
\end{align*}

Now, to estimate the log copula density $\log c_\Sigma(\cdot)$, we provide a reparameterization method for the copula samples $q \sim C_\Sigma(\Phi_1(q_1),...\Phi(q_d))$. As suggested by ~\citet{kingma2014auto, hoffman2013stochastic}, reparameterization is needed as it gives a differentiable, low-variance estimator of the objective function. Here, we parameterize the copula samples with a deterministic function with respect to the Cholesky factor $L$ of its covariance matrix $\Sigma$. We use the fact that for any multivariate Gaussian random variables, a linear transformation of them is also a multivariate Gaussian random variable. Formally, if $X \sim \mathcal{N}(\mu, \Sigma)$, and $Y = AX$, then we must have $Y \sim \mathcal{N}(A\mu, A\Sigma A^T)$. Hence, for a Gaussian copula with the form $c_\Sigma = \mathcal{N}(0, \Sigma)$, we can reparameterize its samples $q$ by,
\begin{align*}
    \epsilon & \sim \mathcal{N}(0, I)\\
    q & = L \cdot \epsilon
\end{align*}
It is easy to see that $q = L \cdot \epsilon \sim \mathcal{N}(0, LI^-L^T=\Sigma)$ is indeed a sample from the Gaussian copula model. This is the standard way of sampling from Gaussian distribution with covariance matrix $LL^T$. To ensure \textbf{numerical stability} of the above reparameterization and to ensure that the covariance $\Sigma=LL^T$ is positive definite, we provide the following algorithm to parameterize $L$. 

\begin{algorithm}
\caption{Neural reparameterization of Copula: Cholesky approach}
\begin{algorithmic}
\STATE h = LSTM(x)\\
\item $w$ = ReLU($W_1 \cdot h + b_1$)
\item $a$ = Tanh($W_2 \cdot h + b_2$)
\item $\Sigma$ = $w \cdot I + aa^T$
\item L = CholeskyFactorization($\Sigma$)
\end{algorithmic}
\end{algorithm}

\begin{table*}[t]
\small
    \centering
    \begin{tabular}{l}
    \hline

    the company said it will be sold to the company 's promotional programs and \_UNK\\
the company also said it will sell \$  n million of soap eggs turning millions of dollars \\

the company said it will be \_UNK by the company 's \_UNK division n\\

the company said it would n't comment on the suit and its reorganization plan \\
    \hline
  mr . \_UNK said the company 's \_UNK group is considering a \_UNK standstill agreement with the company  
\\
  traders said that the stock market plunge is a \_UNK of the market 's rebound in the dow jones industrial average \\
  one trader of \_UNK said the market is skeptical that the market is n't \_UNK by the end of the session\\
   the company said it expects to be fully operational by the company 's latest recapitalization\\
    \hline
    \\ 
    \hline

    i was excited to try this place out for the first time and i was disappointed .\\ the food was good and the food
    a few weeks ago , i was in the mood for a \_UNK of the \_UNK\\
    i love this place . i ' ve been here a few times and i ' m not sure why i ' ve been\\
    this place is really good . i ' ve been to the other location many times and it 's very good . \\
    \hline
    i had a great time here . i was n't sure what i was expecting . i had the \_UNK and the\\
    i have been here a few times and have been here several times . the food is good , but the food is good\\
    this place is a great place to go for lunch . i had the chicken and waffles . i had the chicken and the \_UNK\\
    i really like this place . i love the atmosphere and the food is great . the food is always good . \\
    \hline
    \end{tabular}
    
    \caption{Qualitative comparison between VAE and our proposed approach. First row: PTB samples generated from prior $p(\bold z)$ by VAE (\textit{upper half}) and copula-VAE (\textit{lower half}). Second row: Yelp samples generated from prior $p(\bold z)$ by VAE (\textit{upper half}) and copula-VAE (\textit{lower half}).}
    
    \label{sample1}
    % \vspace{-2ex}
\end{table*}

In Algorithm 1, we first parameterize the covariance matrix and then perform a Cholesky factorization~\cite{chen2008algorithm} to get the Cholesky factor $L$. The covariance matrix $\Sigma = w\cdot I + aa^T$ formed in this way is guaranteed to be positive definite. It is worth noting that we do not sample the latent codes from Gaussian copula. In fact, $\bold z$ still comes from the independent Gaussian distribution. Rather, we get sample $q$ from Gaussian copula $C_\Sigma$ so that we can compute the log copula density term in the following, which will then be used as a regularization term during training, in order to force the learned $\bold z$ to respect the dependencies among individual dimensions.

Now, to calculate the log copula density, we only need to do,
\begin{align*}
    \log c_\Sigma &= \sum_i^d \log \sigma_i - 1/2 \log |\Sigma| + 1/2 q^TMq
\end{align*}
where $M = diag(\Sigma^-) - \Sigma^-$.

To make sure that our model maintains the dependency structure of the latent codes, we seek to maximize both the ELBO and the joint log posterior likelihood $\log q(\bold z|\bold x)$ during the training. In other words, we maximize the following modified ELBO,
\begin{align*}
    \mathcal{L'} = \mathcal{L} + \lambda(\log c_\Sigma(\cdot) + \sum_i^d\log q_\phi(z_i|\bold x))
\end{align*}
where $\mathcal{L}$ is the original ELBO. $\lambda$ is the weight of log density of the joint posterior. It controls how good the model is at maintaining the dependency relationships of latent codes. The reparameterization tricks both for $\bold z$ and $q$ makes the above objective fully differentiable with respect to $\mu, \sigma^2, \Sigma$. Maximizing $\mathcal{L'}$ will then maximize the log input likelihood $\log p(\bold x)$ and the joint posterior log-likelihood $\log q(\bold z|\bold x)$. If the posterior collapses to the prior and has a factorized form, then the joint posterior likelihood will not be maximized since the joint posterior is unlikely factorized. Therefore, maximizing the joint posterior log-likelihood along with ELBO forces the model to generate readable texts while also considering the dependency structure of the true posterior distribution, which is never factorized. 

\subsection{Evidence Lower Bound}
Another interpretation can be seen by taking a look at the prior. If we compose the copula density with the prior, then, like the non-factorized posterior, we can get the non-factorized prior,
\begin{align*}
    \log p_\theta(\bold z) = \log c_\Sigma(u_1,...,u_d)+\sum_i^d \log p_\theta(z_i)
\end{align*}
And the corresponding ELBO is,
\begin{align*}
\mathcal{L}(\bold \theta; \bold \phi; \bold x)
& = \E_{q(\bold x)}[\E_{q_\phi(\bold z| \bold x)}[\log p_\theta(\bold x| \bold z)] \\
& \qquad - KL(q_\phi(\bold z|\bold x)||p_\theta(\bold z))]\\
& = \E_{q(\bold x)}[\E_{q_\phi(\bold z| \bold x)}[\log p_\theta(\bold x| \bold z)] \\
& \qquad - \E_{q_\phi(\bold z|\bold x)}[\log q_\phi(\bold z|\bold x) - \log p(\bold z)] \\
& \qquad \qquad + \E_{q_\phi(\bold z|\bold x)}[\log c_\Sigma(u_1,...,u_d)]]
\end{align*}
Like Normalizing flow~\cite{rezende2015variational}, maximizing the log copula density will then learns a more flexible prior other than a standard Gaussian. The dependency among each $z_i$ is then restored since the KL term will push the posterior to this more complex prior.

We argue that relieving the Mean-field assumption by maintaining the dependency structure can avert the posterior collapse problem. As shown in Figure~\ref{fig:stage}, during the training stage of original VAE, if kl-annealing~\cite{bowman2015generating} is used, the model first seeks to maximize the expectation $\E_{q(\bold z|\bold x)}[p(\bold x|\bold z)]$. Then, since $q(\bold z|\bold x)$ can never reach to the true $p(\bold z|\bold x)$, $q$ will reach to a boundary and then the expectation can no longer increase. During this stage, the model starts to maximize the ELBO by minimizing the KL divergence. Since the expectation is maximized and can no longer leverage KL, the posterior will collapse to the prior and there is not sufficient gradient to move it away since ELBO is already maximized. On the other hand, if we introduce a copula model to help maintain the dependency structure of the true posterior by maximizing the joint posterior likelihood, then, in the ideal case, the variational family can approximate distributions of any forms since it is now not restricted to be factorized, and therefore it is more likely for $q$ in Figure~\ref{fig:copula} to be closer to the true posterior $p$. In this case, the $\E_{q(\bold z|\bold x)}[p(\bold x|\bold z)]$ can be higher since now we have latent codes sampled from a more accurate posterior, and then this expectation will be able to leverage the decrease of KL even in the final training stage.

\section{Experimental Results}

\subsection{Datasets}

\begin{table}[h]
\small
    \begin{center}
    \begin{tabular}{l|c c c c}
    \hline  \bf Data & \bf Train & \bf Valid & \bf Test & \bf Vocab \\ \hline
        Yelp13 & 62522 & 7773 & 8671 & 15K\\
        PTB & 42068 & 3370 & 3761 & 10K \\
        Yahoo & 100K & 10K & 10K & 20K\\
    \hline
    \end{tabular}
    \end{center}
    \caption{Size and vocabulary size for each dataset.}
    \label{tab:data}
    \vspace{-2ex}
\end{table}
In the paper, we use Penn Tree~\cite{marcus1993building}, Yahoo Answers~\cite{xu2018spherical, yang2017improved}, and Yelp 13 reviews~\cite{xu2016cached} to test our model performance over variational language modeling tasks. We use these three large datasets as they are widely used in all other variational language modeling approaches~\cite{bowman2015generating,yang2017improved, xu2018spherical, xiao2018dirichlet, he2018lagging, kim2018semi}. Table~\ref{tab:data} shows the statistics, vocabulary size, and number of samples in Train/Validation/Test for each dataset.

\begin{figure*}[t!]
\small
    \centering
    \begin{subfigure}{.4\textwidth}
        \centering
        \includegraphics[width=0.9\linewidth]{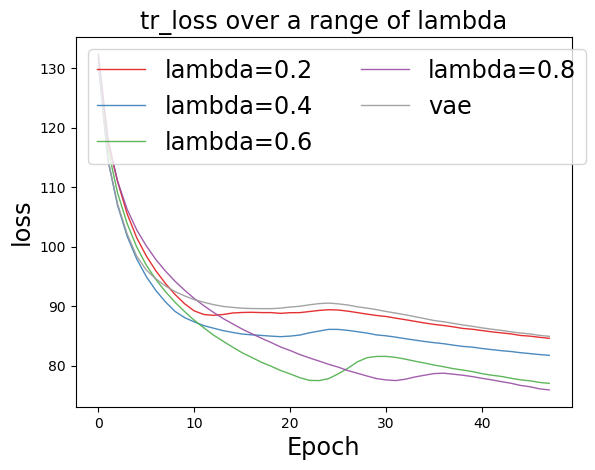}
        \caption{train loss}
    \end{subfigure}
    % \begin{subfigure}{.4\textwidth}
    %     \centering
    %     \includegraphics[width=0.9\linewidth]{tr_cho_ppl.png}
    %     \caption{train ppl}
    % \end{subfigure}
    \begin{subfigure}{.4\textwidth}
        \centering
        \includegraphics[width=0.9\linewidth]{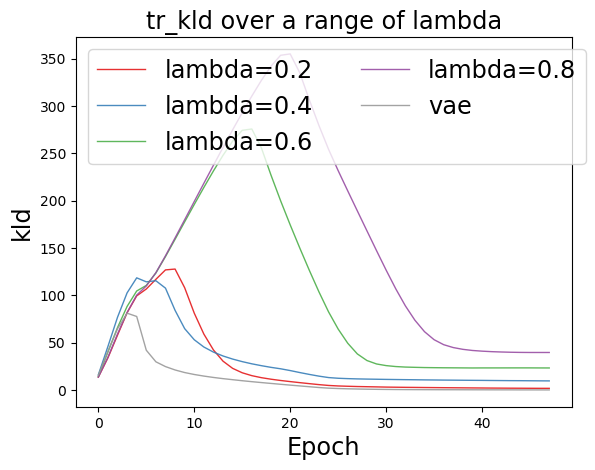}
        \caption{train kld}
    \end{subfigure}
    \begin{subfigure}{.4\textwidth}
        \centering
        \includegraphics[width=0.9\linewidth]{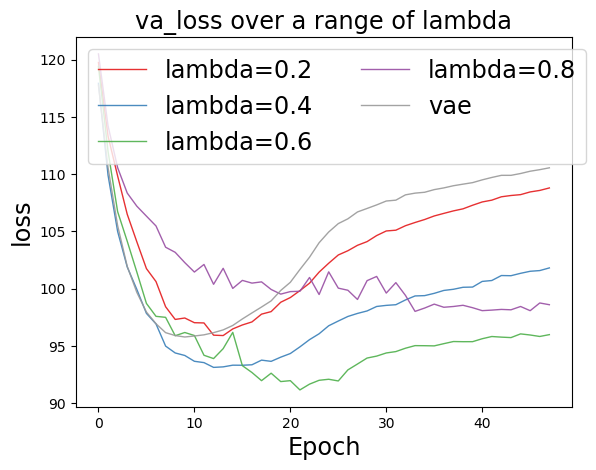}
        \caption{validation loss}
    \end{subfigure}
    \begin{subfigure}{.4\textwidth}
        \centering
        \includegraphics[width=0.9\linewidth]{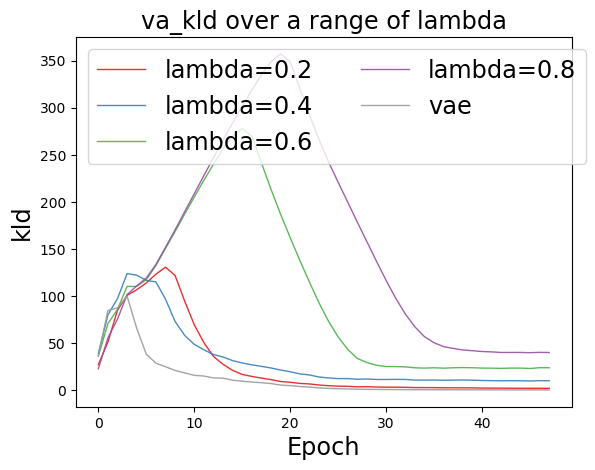}
        \caption{validation kld}
    \end{subfigure}

    \caption{Training and Validation KL divergence and sequential loss for PTB using Cholesky Neural Copula with different copula density weights $\lambda$. It is obvious that the optimal $\lambda$ plays a huge role in alleviating the posterior collapse problem and does not result in overfitting as VAE does. Models are trained using 1-layer LSTM with 200 hidden units for Encoder/Decoder, in which the embedding size is 200 and the latent dimension is 32.}
    \label{fig:hyperplot1}
\end{figure*}

\begin{figure}[t!]

    \centering
   
        \includegraphics[width=0.9\linewidth]{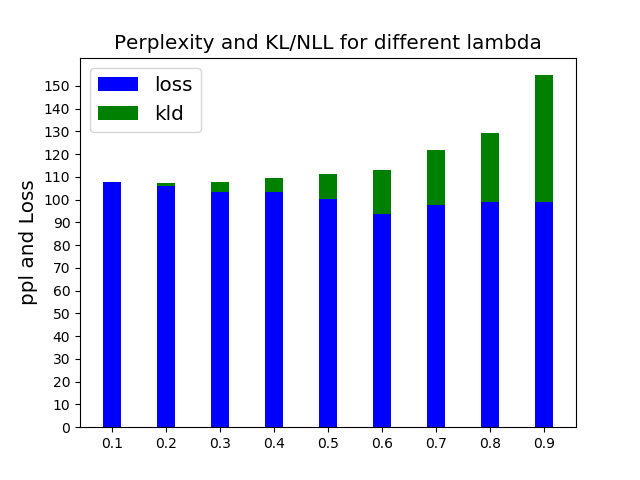}
        
    \caption{Reconstruction loss, KL divergence, and PPL (sum of loss and KL) on PTB. When we gradually increase $\lambda$, the KL divergence increases and the test reconstruction will decrease.}
    \label{fig:s1}
\end{figure}

\subsection{Experimental Setup}
We set up a similar experimental condition as in ~\cite{bowman2015generating, xiao2018dirichlet, yang2017improved, xu2018spherical}. We use LSTM as our encoder-decoder model, where the number of hidden units for each hidden state is set to 512. The word embedding size is 512. And, the number of dimension for latent codes is set to 32. For both encoder and decoder, we use a \textit{dropout} layer for the initial input, whose dropout rate is $\alpha=0.5$. Then, for inference, we pass the final hidden state to a linear layer following a \textit{Batch Normalizing}~\cite{ioffe2015batch} layer to get reparameterized samples from $\prod_i q(z_i|x)$ and from Gaussian copula $C_\Sigma$. For the training stage, the maximum vocabulary size for all inputs are set to 20000, and the maximum sequence length is set to 200. Batch size is set to 32, and we train for 30 epochs for each dataset, where we use the \textit{Adam stochastic optimization}~\cite{kingma2014adam} whose learning rate is $r=10^{-3}$.

We use \textit{kl annealing}~\cite{bowman2015generating} during training. We also observe that the weight of \textit{log copula} density is the most important factor which determines whether our model avoids the posterior collapse problem. We hyper tuned this parameter in order to find the optimal one.

\subsection{Language Modeling Comparison}

\begin{table}
\small
    \begin{tabular}{l l l l}
    \hline \bf Model \qquad & \bf NLL & \bf KL & \bf PPL\\
    \hline 
        LSTM-LM~\cite{yang2017improved} & 116.2 & - & 104.2  \\
        VAE~\cite{bowman2015generating} & 105.2 & 1.74 & 121\\
        vmf-VAE~\cite{xu2018spherical} & 96.0 & 5.7 & 79.6 \\
        VAE-NF & 96.8 & 0.87 & 82.9\\
        non-diag-VAE & 105.1 & 4.9 & 121.0\\
        \bf copula-VAE(cho) $\lambda=0.4$ &\bf 92.2 &\bf 7.3 &\bf 67.2 \\
        \hline
    \end{tabular}
    \caption{Variational language modeling on PTB}
    \label{tab:language_ptb}

    \begin{tabular}{l l l l}
    \hline \bf Model \qquad & \bf NLL & \bf KL & \bf PPL\\
    \hline 
        VAE~\cite{bowman2015generating} & 197.1 & 0.03 & 58.2\\
        vmf-VAE~\cite{xu2018spherical} & 198.0 & 6.4 & 59.3 \\
        VAE-NF & 200.4& 0.1 & 62.5\\
        non-diag-VAE & 198.3 & 4.7 & 59.7\\
        \bf copula-VAE(cho) $\lambda=0.5$ &\bf 187.7 &\bf 10.0 &\bf 48.0 \\
        \hline
    \end{tabular}
    \caption{Variational language modeling on Yelp Reviews 13}
    \label{tab:language_yelp}
        
    \begin{tabular}{l l l l}
    \hline \bf Model \qquad & \bf NLL & \bf KL & \bf PPL\\
    \hline 
        VAE~\cite{bowman2015generating} & 351.6 & 0.3 & 81.6\\
        vmf-VAE~\cite{xu2018spherical} & 359.3 & 17.9 & 89.9 \\
        VAE-NF & 353.8 & 0.1 & 83.0\\
        lagging-VAE~\cite{he2018lagging} & \bf 326.6 & \bf 6.7 & \bf 64.9 \\
        non-diag-VAE & 352.2 & 5.7 & 82.3\\
        \bf copula-VAE(cho) $\lambda=0.5$ & 344.2 & 15.4 & 74.4 \\
        \hline
    \end{tabular}
    \caption{Variational language modeling on Yahoo}
    \label{tab:language_yahoo}
\end{table}

\subsubsection{Comparison with other Variational models}

% \begin{table}[t]
% \small
%     \begin{tabular}{l l l l}
%     \hline \bf Model \qquad & \bf NLL & \bf KL & \bf PPL\\
%     \hline 
%         WAE~\cite{wang2019rnfwae} & 94.0 & 24.4& 37.2\\
%         \bf copula-VAE(cho) $\lambda=0.5$ &\bf 60.0 &\bf 26.6 &\bf 39.7 \\
%         \hline
%     \end{tabular}
%     \caption{Comparison with WAE on PTB}
%     \label{tab:sota1}
%     \begin{tabular}{l l l l}
%     \hline \bf Model \qquad & \bf NLL & \bf KL & \bf PPL\\
%     \hline 
%         WAE~\cite{wang2019rnfwae} & 183.6 & 20.6& 38.2\\
%         \bf copula-VAE(cho) $\lambda=0.5$ &\bf 185.8 &\bf 8.7 &\bf 45.2 \\
%         \hline
%     \end{tabular}
%     \caption{Comparison with WAE on Yelp}
%     \label{tab:sota2}
    
% \end{table}
We compare the variational language modeling results over three datasets. We show the results for Negative log-likelihood (NLL), KL divergence, and sample \text{perplexity (PPL)} for each model on these datasets. NLL is approximated by the \textit{evidence lower bound}.

First, we observe that kl-annealing does not help alleviate the posterior collapse problem when it comes to larger datasets such as Yelp, but the problem is solved if we can maintain the latent code's dependencies by maximizing the copula likelihood when we maximize the ELBO. We also observe that the weight $\lambda$ of log copula density affects results dramatically. All $\lambda$ produce competitive results compared with other methods. Here, we provide the numbers for those weights $\lambda$ that produce the lowest PPL. For PTB, copula-VAE achieves the lowest sample perplexity, best NLL approximation, and do not result in posterior collapse when $\lambda=0.4$. For Yelp, the lowest sample perplexity is achieved when $\lambda=0.5$.

We also compare with VAE models trained with normalizing flows~\cite{rezende2015variational}. We observed that our model is superior to VAE based on flows. It is worth noting that Wasserstein Autoencoder trained with Normalizing flow~\cite{wang2019riemannian} achieves the lowest PPL 66 on PTB, and 41 on Yelp. However, the problem of designing flexible normalizing flow is orthogonal to our research.

\subsubsection{Generation}
Table~\ref{sample1} presents the results of text generation task. We first randomly sample $\bold z$ from $p(\bold z)$, and then feed it into the decoder $p(\bold x| \bold z)$ to generate text using greedy decoding. We can tell whether a model suffers from posterior collapse by examining the diversity of the generated sentences. The original VAE tends to generate the same type of sequences for different $\bold z$. This is very obvious in PTB where the posterior of the original VAE collapse to the prior completely. Copula-VAE, however, does not have this kind of issue and can always generate a diverse set of texts.

\subsection{Hyperparameter-Tuning: Copula weights play a huge role in the training of VAE}

In this section, we investigate the influence of log copula weight $\lambda$ over training. From Figure~\ref{fig:s1}, we observe that our model performance is very sensitive to the value of $\lambda$. We can see that when $\lambda$ is small, the log copula density contributes a small part to the objective, and therefore does not help to maintain dependencies over latent codes. In this case, the model performs like the original VAE, where the KL divergence becomes zero at the end. When we increase $\lambda$, test KL becomes larger and test reconstruction loss becomes smaller. 

This phenomenon is also observed in validation datasets, as shown in Figure~\ref{fig:hyperplot1}. The training PPL is monotonically decreasing in general. However, when $\lambda$ is small and the dependency relationships over latent codes are lost, the model quickly overfits, as the KL divergence quickly becomes zero and the validation loss starts to increase. This further confirms what we showed in Figure~\ref{fig:stage}. For original VAE models, the model first maximizes $\E_{q(\bold z|\bold x)}[p(\bold x|\bold z)]$ which results in the decrease of both train and validation loss. Then, as $q(\bold z|\bold x)$ can never match to the true posterior, $\E_{q(\bold z|\bold x)}[p(\bold x|\bold z)]$ reaches to its ceiling which then results in the decrease of KL as it is needed to maximize the ELBO. During this stage, the LSTM decoder starts to learn how to generate texts with standard Gaussian latent variables which then causes the increase of validation loss. On the other hand, if we gradually increase the contribution of copula density by increasing the $\lambda$, the model is able to maintain the dependencies of latent codes and hence the structure of the true posterior. In this case, $\E_{q(\bold z|\bold x)}[p(\bold x|\bold z)]$ will be much higher and will leverage the decrease of KL. In this case, the decoder is forced to generate texts from non-standard Gaussian latent codes. Therefore, the validation loss also decreases monotonically in general.

One major drawback of our model is the amount of training time, which is 5 times longer than the original VAE method.
In terms of performance, copula-VAE achieves the lowest reconstruction loss when $\lambda=0.6$. It is clear that from Figure~\ref{fig:s1} that increasing $\lambda$ will result in larger KL divergence.

\section{Conclusion}
In this paper, we introduce \textbf{Copula-VAE} with Cholesky reparameterization method for Gaussian Copula. This approach averts \textbf{Posterior Collapse} by using Gaussian copula to maintain the dependency structure of the true posterior. Our results show that Copula-VAE significantly improves the language modeling results of other VAEs. 
% \textcolor{red}{One table here: compare the numbers}

% \subsection{Samples from copula}

% \section{Sentiment Analysis}

% \textcolor{red}{One table here: Classification}

% \section{Text Generation}

% \textcolor{red}{Several tables of texts here}

\bibliography{acl2019}
\bibliographystyle{acl_natbib}

% \section{Supplemental Material}
% \label{sec:supplemental}

% \begin{figure*}[h!]
% \small
%     \centering
%     \begin{subfigure}{.3\textwidth}
%         \centering
%         \includegraphics[width=0.9\linewidth]{tr_ldl_ppl.png}
%         \caption{train ppl, ldl}
%     \end{subfigure}
%     \begin{subfigure}{.3\textwidth}
%         \centering
%         \includegraphics[width=0.9\linewidth]{tr_ldl_kld.png}
%         \caption{train kld, ldl}
%     \end{subfigure}
%     \begin{subfigure}{.3\textwidth}
%         \centering
%         \includegraphics[width=0.9\linewidth]{tr_ldl_loss.png}
%         \caption{train loss, ldl}
%     \end{subfigure}\\
    
%     \begin{subfigure}{.3\textwidth}
%         \centering
%         \includegraphics[width=0.9\linewidth]{va_ldl_ppl.png}
%         \caption{validation ppl, ldl}
%     \end{subfigure}
%     \begin{subfigure}{.3\textwidth}
%         \centering
%         \includegraphics[width=0.9\linewidth]{va_ldl_kld.png}
%         \caption{validation kld, ldl}
%     \end{subfigure}
%     \begin{subfigure}{.3\textwidth}
%         \centering
%         \includegraphics[width=0.9\linewidth]{va_ldl_loss.png}
%         \caption{validation loss, ldl}
%     \end{subfigure}
%     \caption{Training and Validation perplexity, KL divergence, and sequential loss for PTB using $LDL^T$ Neural Copula with different copula density weights $\lambda$. \william{I would put this in the supplementary materials. Fonts are too small, and remove the gray background color. Likewise for Figure 6.}}
%     \label{fig:hyperplot2}
% \end{figure*}

\end{document}